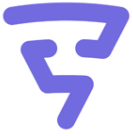

# Developing an Open Conversational Speech Corpus for the Isan Language
# การสร้างคลังข้อมูลเสียงพูดภาษาอีสาน

Typhoon Team, SCB 10X
contact@opentyphoon.ai

# Abstract

This paper introduces the development of the first open conversational speech dataset for the Isan language, the most widely spoken regional dialect in Thailand. Unlike existing speech corpora that are primarily based on read speech, this dataset consists of natural speech, thereby capturing authentic linguistic phenomena such as colloquials, spontaneous prosody, disfluencies, and frequent code-switching with central Thai.

A key challenge in building this resource lies in the lack of a standardized orthography for Isan. Current writing practices vary considerably, due to the different lexical tones between Thai and Isan. This variability complicates the design of transcription guidelines and poses questions regarding consistency, usability, and linguistic authenticity. To address these issues, we establish practical transcription protocols that balance the need for representational accuracy with the requirements of computational processing.

By releasing this dataset as an open resource, we aim to contribute to inclusive AI development, support research on underrepresented languages, and provide a basis for addressing the linguistic and technical challenges inherent in modeling conversational speech.

# 1    ที่มาและความสำคัญของปัญหา

ปัจจุบัน กระบวนการวิจัยและพัฒนาปัญญาประดิษฐ์ด้านเสียงสำหรับภาษาไทยต่างมุ่งเน้นการประมวลผลและการรองรับภาษาไทยกลางเป็นหลัก ในขณะที่ชุดข้อมูลเสียงภาษาไทยถิ่นอื่น ๆ ยังคงมีจำกัด ทำให้ขอบเขตของการพัฒนาปัญญาประดิษฐ์ที่สามารถตอบสนองความต้องการของผู้ใช้ภาษาถิ่นจึงยังคงถูกจำกัดอยู่ในวงแคบและนำไปสู่ความเหลื่อมล้ำในการเข้าถึงเทคโนโลยีดิจิทัล ด้วยเหตุนี้ ผู้จัดทำจึงเล็งเห็นถึงความสำคัญของการแก้ไขข้อจำกัดนี้ และมุ่งเน้นการเริ่มต้นพัฒนาและรวบรวมชุดข้อมูลเสียงภาษาถิ่น เพื่อเป็นรากฐานสำคัญในการสร้างสรรค์ระบบปัญญาประดิษฐ์ที่ครอบคลุมผู้ใช้ภาษาทุกกลุ่มทั่วประเทศ

ผู้จัดทำพิจารณาทำชุดข้อมูลภาษาอีสานเป็นภาษาแรก เนื่องจากเป็นภาษาถิ่นที่มีผู้พูดเยอะที่สุดในประเทศไทย (International Convention on the Elimination of All Forms of Racial Discrimination, 2011) รายงานฉบับนี้จัดทำขึ้นเพื่ออธิบายแนวคิดเบื้องหลังในการสร้างชุดข้อมูล โดยมีเนื้อหาครอบคลุมตั้งแต่การจัดกลุ่มสำเนียงภาษาอีสาน กระบวนการเก็บข้อมูลเสียง แนวทางการสะกดภาษาอีสาน แนวทางการถอดข้อความเสียงเป็นตัวอักษร (transcription) และแนวทางการถ่ายถอดเสียงเพื่อสร้างพจนานุกรมคำอ่าน (phonetic dictionary) พร้อมชุดข้อมูลเสียงภาษาอีสานและพจนานุกรมคำอ่านภาษาอีสาน เพื่อเป็นรากฐานและแนวทางในการพัฒนาเทคโนโลยีเสียงภาษาถิ่นอื่น ๆ ของประเทศไทยในอนาคตต่อไป

## 2  ชุดข้อมูลที่มีอยู่ในปัจจุบัน

ชุดข้อมูลภาษาถิ่นอีสานในปัจจุบันมีอยู่เพียงสองชุด ชุดข้อมูลชุดแรกคือชุดข้อมูลของ Suwanbandit et al. (2023) ซึ่งรวบรวมข้อมูลเสียงพูดที่บันทึกจากผู้พูดภาษาโคราช เก็บข้อมูลโดยให้อ่านข้อความภาษาโคราชที่แปลมาจากภาษาไทยกลาง ชุดข้อมูลที่สองคือ Thatphithakkul, K. Thangthai and V. Chunwijitra (2024) ซึ่งรวบรวมข้อมูลเสียงพูดที่บันทึกจากผู้พูดภาษาอีสานในจังหวัดขอนแก่น เก็บข้อมูลโดยให้ผู้บอกภาษาอ่านข้อความภาษาอีสานที่คาดว่าน่าจะแปลมาจากภาษาไทยกลางเนื่องจากทุกภาษาในชุดข้อมูลมีเนื้อความเดียวกัน

อย่างไรก็ตาม ชุดข้อมูลทั้งสองข้อมูลนี้ยังมีข้อจำกัดอยู่หลายประการ ประการที่หนึ่งคือข้อจำกัดทางภาษาถิ่นที่เลือก สำหรับ Suwanbandit et al. (2023) ภาษาโคราชอาจไม่พิจารณานับเป็นภาษาถิ่นอีสาน เนื่องจากทั้งผู้พูดภาษาโคราชและผู้พูดภาษาอีสานสำเนียงอื่น ๆ มิได้มีภาพจำว่าภาษาโคราชเป็นภาษาที่ใช้ในภาคอีสาน หากแต่เป็นภาษาถิ่นของจังหวัดนครราชสีมาเท่านั้น (ถาวร สุบงกช, 2526) ในขณะที่ชุดข้อมูลของ Thatphithakkul, K. Thangthai and V. Chunwijitra (2024) นั้นมุ่งไปที่ภาษาอีสานขอนแก่นเพียงจังหวัดเดียวและไม่ได้ครอบคลุมพื้นที่อื่น ๆ ของภาคอีสาน จึงอาจไม่ครอบคลุมคำศัพท์หรือสำเนียงในจังหวัดอื่นในภูมิภาค ด้วยเหตุนี้ ชุดข้อมูลทั้งสองชุดจึงยังไม่อาจนับเป็นชุดข้อมูลที่เป็นตัวแทนของภาษาอีสานได้อย่างแท้จริง

ประการที่สองคือข้อจำกัดทางวิธีการพูด เนื่องจากผู้พูดจะออกเสียงแตกต่างกันระหว่างการอ่านออกเสียงจากข้อความที่เห็น (narrative) และเมื่อสนทนา (conversational) หรือพูดจากความรู้สึกนึกคิดของตนเอง ยกตัวอย่างเช่น การออกเสียงคำว่า 'ครับ' สามารถมีการลากเสียง การกร่อนเสียง หรือการขึ้นลงของน้ำเสียง กลายเป็น "คับ" "ขับ" "คาบ" "ค้าบ" "คร้าบ" นอกจากนี้ ในการพูดสนทนา ผู้พูดมักจะมีคำแสดงความลังเลปรากฏร่วมด้วย เช่น "อืม" "เอ่อ" "อ่า" ด้วยเหตุนี้ เมื่อชุดข้อมูลทั้งสองชุดเก็บข้อมูลเสียงด้วยการอ่านหรือแบบ narrative ชุดข้อมูลที่ออกมาอาจขาดปรากฏการณ์ทางเสียงที่พบได้ในการพูดแบบสนทนาหรือแบบ conversational อันจะส่งผลให้โมเดลที่เทรนนั้นไม่สามารถจับคำพูดแบบ conversational ได้อย่างแม่นยำ

ประการที่สามคือข้อจำกัดของสิ่งเร้า (stimuli) สำหรับการบันทึกเสียงได้มาจากการแปล โดยในขั้นตอนของการแปล ผู้แปลมีโอกาสได้รับอิทธิพลภาษาต้นทางสูง และทำให้ข้อความในภาษาปลายทางแตกต่างจากธรรมชาติในการใช้ภาษาจริงของผู้พูด ด้วยเหตุนี้ เมื่อทั้งชุดข้อมูลทั้งสองชุดเก็บข้อมูลเสียงโดยใช้สิ่งเร้าที่แปลมาจากข้อความจากภาษาอื่น ชุดข้อมูลจึงอาจไม่ตรงกับการใช้ภาษาจริงของเจ้าของภาษา เพราะอาจจะขาดคำหรือไวยากรณ์แบบภาษาถิ่นที่ผู้พูดใช้ตามปกติเมื่อไม่ได้รับอิทธิพลจากภาษาอื่น ๆ อันจะส่งผลให้โมเดลที่เทรนนั้นขาดการเรียนรู้ศัพท์และไวยากรณ์ภาษาถิ่นไปด้วย

ประการสุดท้ายคือข้อจำกัดของการสะกดคำ โดยสามารถแบ่งประเด็นการสะกดได้สามประเด็น คือ

**ประเด็นที่ 1 การสะกดขัดกับอักขรวิธีของภาษาไทย** ข้อความถอดมีการสะกดที่ขัดกับอักขรวิธีของภาษาไทย เช่น การสะกด "สังข์หยด" เป็น "สังหย่ด" (Suwanbandit et al., 2023) และการสะกด "แหวกลึก" เป็น "แหวกเลิ ็ก" (Thatphithakkul, K. Thangthai & V. Chunwijitra, 2024) ซึ่งขัดอักขรวิธีภาษาไทยกลางที่จะไม่ใส่ไม้จัตวากับคำตาย และไม่ใส่ไม้ไต่คู้กับรูปสระที่เขียนด้านบน

**ประเด็นที่ 2 การสะกดไม่คุ้นชินสำหรับผู้ใช้ภาษา** ข้อความถอดไม่คุ้นชินสำหรับผู้ใช้ภาษาแม้จะใกล้เคียงกับเสียงที่ออกจริง เช่น การสะกด "รหัส" เป็น "รหั้ส" (Suwanbandit et al., 2023) และการสะกด "แยกแยะบทบาท" เป็น "แญกแญะบทบาท" (Thatphithakkul, K. Thangthai & V. Chunwijitra, 2024) เนื่องจากผู้ใช้ภาษาอีสานในประเทศไทยได้รับอิทธิพลจากภาษาไทยกลาง (McCargo & Hongladarom, 2004) ดังนั้นก็จะคุ้นชินกับการสะกดแบบไทยกลางว่า "รหัส" และ "แยกแยะ" ตามลำดับมากกว่า

**ประเด็นที่ 3 การสะกดไม่สม่ำเสมอ** ข้อความถอดสะกดคำเดียวกันไม่เหมือนกัน เช่น สะกด "รหัส" เป็นทั้ง "ระหัส" และ "รหั้ส" (Suwanbandit et al. (2023) และสะกด "เรื่อง" เป็นทั้ง "เลื่อง" และ "เรื่อง" (Thatphithakkul, K. Thangthai & V. Chunwijitra, 2024)

ด้วยข้อจำกัดทั้งสามประการนี้ ชุดข้อมูลที่ออกมาจึงมีความไม่สม่ำเสมอ อันจะส่งผลให้โมเดลที่เทรนนั้นสะกดข้อความที่ถอดไม่สม่ำเสมอ และอาจมีข้อความที่ผู้ใช้ภาษาไม่ใช้จริงปะปนเข้าไปด้วย

งานนี้จึงมุ่งเน้นที่จะแก้ปัญหาเหล่านี้ โดยเริ่มตั้งแต่นิยามภาษาถิ่นอีสานโดยใช้หลักเกณฑ์ทางภาษาศาสตร์ ออกแบบการเก็บบันทึกข้อมูลเสียงเพื่อให้ได้เสียงพูดแบบสนทนาที่เป็นธรรมชาติ และกำหนดหลักเกณฑ์การสะกดคำและการถอดความเพื่อความสม่ำเสมอและเป็นบรรทัดฐานเดียวกันสำหรับผู้ถอดความหลายคน โดยจะอธิบายกระบวนการเหล่านี้ในส่วนต่อไป

# 3 คุณสมบัติของภาษาอีสาน

## 3.1 นิยามของภาษาอีสานและสำเนียงต่าง ๆ

เพื่อลดข้อจำกัดด้านความไม่หลากหลายของสำเนียงในภาษาอีสาน กระบวนการแรกก่อนการเก็บข้อมูลภาษาอีสานคือการนิยามภาษาอีสานและการจำแนกสำเนียงต่าง ๆ เนื่องจากในความเข้าใจของคนทั่วไป ภาษาอีสานคือภาษาใช้ในจังหวัดที่ตั้งอยู่ในภาคตะวันออกเฉียงเหนือ ซึ่งนิยามนี้จะทำให้ภาษาอีสานไม่ได้มีสำเนียงเดียว ดังภาพที่ 1 แสดงขอบเขตของภาษาถิ่นต่าง ๆ ในภาคตะวันออกเฉียงเหนือ การเลือกสำเนียงหลักในการทำชุดข้อมูลจึงจำเป็น โดยมีจุดมุ่งหมายเพื่อให้ครอบคลุมการใช้งานของผู้ใช้ภาษาส่วนใหญ่และเป็นนิยามของสำเนียงที่ตรงกับความเข้าใจของคนทั่วไป

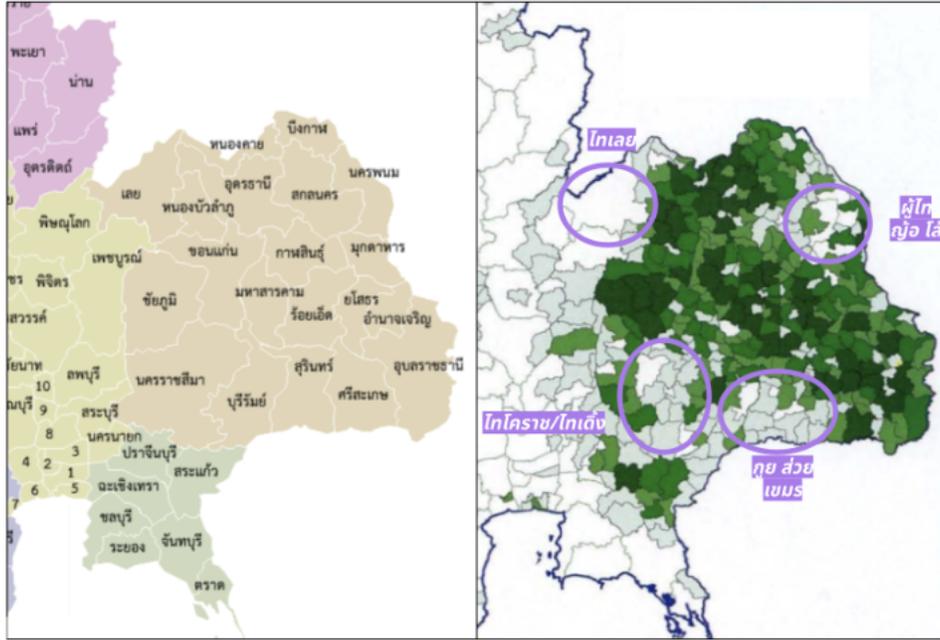

**ภาพที่ 1**. ซ้ายมือแสดงแผนที่การปกครองและขวามือแสดงแผนที่การกระจายตัวของภาษาอีสาน (สุวิไล เปรมศรีรัตน์ และคณะ, 2547)

ความสำคัญของการเลือกสำเนียงหลักในที่นี้เปรียบได้กับการเลือกสำเนียงสำหรับพัฒนาโมเดลภาษาไทยกลาง ในภาคกลางของประเทศไทยประกอบไปด้วยหลายสำเนียง เช่น สำเนียงสุพรรณบุรี สำเนียงอยุธยา สำเนียงกรุงเทพ เป็นต้น ทว่าการทำโมเดลหรือสร้างชุดข้อมูลก็จะเลือกสำเนียงกรุงเทพเป็นสำเนียงหลัก เพื่อให้ชุดข้อมูลมีความเป็นเอกภาพและทำให้เกิดการใช้งานกว้างขวางที่สุด ในทางเดียวกัน การเลือกสำเนียงอีสานหลักจึงมีความจำเป็นเพื่อให้บรรลุวัตถุประสงค์เดียวกันนี้

วิธีการจำแนกสำเนียงมีหลายวิธีด้วยกัน เช่น การสืบสร้างตระกูลภาษา การทดสอบความเข้าใจร่วมกัน การพิจารณาทางเสียงและพัฒนาการทางเสียง อย่างไรก็ตาม สำเนียงในภาษาอีสานมีความคล้ายคลึงกันมากในหลายประการ คือ หนึ่ง ภาษาถิ่นอีสานส่วนใหญ่จะใช้คำศัพท์ที่ค่อนข้างคล้ายคลึงกัน และสอง ภาษาถิ่นอีสานส่วนใหญ่มักมีหน่วยเสียงพยัญชนะและหน่วยเสียงสระเดียวกันหรือคล้ายคลึงกันมาก ดังนั้น เกณฑ์ที่ใช้จำแนกภาษาอีสานได้ชัดเจนที่สุด จึงเป็นเกณฑ์ระบบเสียงวรรณยุกต์ โดยใช้เครื่องมือที่เรียกว่า "กล่องวรรณยุกต์" (tone box) เป็นเครื่องมือหลักในการจำแนกภาษาถิ่นย่อย (dialect classification)

เมื่อใช้เกณฑ์ระบบเสียงวรรณยุกต์แล้ว ภาษาอีสานสามารถจำแนกได้เป็น 2 สำเนียง คือ ภาษาอีสานที่ใช้ระบบ 6 วรรณยุกต์และภาษาอีสานที่ไม่ใช้ระบบ 6 วรรณยุกต์ (พิณรัตน์ อัครวัฒนากุล, 2546; ธนานันท์ ตรงดี, 2557) โดยมีรายละเอียดดังนี้

**กลุ่มที่ 1 ภาษาอีสานที่ใช้ระบบ 6 วรรณยุกต์** ใช้พูดในแถบจังหวัดหนองบัวลำภู อุดรธานี ขอนแก่น หนองคาย กาฬสินธุ์ ร้อยเอ็ด ยโสธร มหาสารคาม อุบลราชธานี อำนาจเจริญ มุกดาหาร ศรีสะเกษ บางอำเภอของเลย บางอำเภอของชัยภูมิ บางอำเภอของสกลนคร และบางอำเภอของบุรีรัมย์

**กลุ่มที่ 2 ภาษาอีสานที่ไม่ใช้ระบบ 6 วรรณยุกต์** กระจายอยู่ตามจังหวัดแถบจังหวัดเลย อุตรดิตถ์ ศรีสะเกษ สกลนคร บางอำเภอของอุบลราชธานี และบางพื้นที่ในประเทศลาว

เนื่องด้วยสำเนียงอีสานที่ใช้ระบบ 6 วรรณยุกต์เป็นสำเนียงที่ครอบคลุมพื้นที่อีสานกว้างขวางที่สุด ทั้งยังเป็นสำเนียงที่ปรากฏในสื่อ ด้วยเหตุนี้ ชุดข้อมูลจึงเลือกสำเนียงอีสานที่ใช้ระบบ 6 วรรณยุกต์เป็นสำเนียงหลักของชุดข้อมูลชุดนี้ โดยจากนี้จะเรียกสำเนียงนี้ว่า 'สำเนียงอีสาน 6 วรรณยุกต์'

สิ่งที่ต้องพิจารณาในการจัดทำชุดข้อมูลสำเนียงอีสาน 6 วรรณยุกต์ คือ ชุดข้อมูลเสียงต้องประกอบไปด้วยทั้งเสียงและข้อความ อย่างไรก็ตาม ภาษาอีสานมีสถานะเป็นภาษาพูด ยังมิได้มีการกำหนดอักขรวิธีการสะกดคำ ทำให้เกิดประเด็นที่ต้องพิจารณาในการสะกดคำภาษาอีสาน ประเด็นแรกคือเสียงวรรณยุกต์ที่ไม่ตรงกับภาษาไทยกลาง เนื่องจากภาษาอีสานมีเสียงวรรณยุกต์มากกว่าภาษาไทยกลาง และเสียงวรรณยุกต์อีสานมีทั้งที่เหมือนหรือคล้ายคลึงกับภาษาไทยกลางและต่างจากภาษาไทยกลาง และประเด็นที่สองคือการออกเสียงสระและพยัญชนะในภาษาอีสานบางคำต่างกับภาษาไทยกลาง

ทั้งสองประเด็นที่ต้องพิจารณาจะมีผลต่อการกำหนดอักขรวิธีการสะกดคำ ส่วนต่อไปจะอธิบายทั้งสองประเด็นนี้ในรายละเอียดปลีกย่อยต่อไป

## 3.2 หลักการจำแนกเสียงวรรณยุกต์ภาษาอีสาน 6 วรรณยุกต์

ในทางภาษาศาสตร์ วรรณยุกต์ หมายถึง การขึ้น-ลงของเสียงเฉพาะคำที่มีอิทธิพลต่อความหมาย เช่น ในภาษาไทยกลาง เมื่อพิจารณาว่า "ปา" และ "ป่า" สองคำนี้องค์ประกอบเหมือนกันทั้งหมด (พยัญชนะต้น ป สระ อา และไม่มีตัวสะกด) แต่ว่ามีความหมายแตกต่างกันเพราะมีการขึ้นเสียง-สูงต่ำต่างกัน ดังนั้น จึงสรุปได้ว่า "ปา" และ "ป่า" มีคนละเสียงวรรณยุกต์

กล่องวรรณยุกต์ (tone box) (Gedney, 1972) เป็นเครื่องมือที่ใช้ในการตรวจสอบเสียงวรรณยุกต์ว่ามีกี่หน่วยเสียงในภาษาและเงื่อนไขการเกิดแต่ละวรรณยุกต์ กล่องวรรณยุกต์เกิดขึ้นเนื่องจากข้อค้นพบที่ว่าวรรณยุกต์ภาษาตระกูลไทขึ้นอยู่กับการกำกับรูปวรรณยุกต์เอกและโทและพยัญชนะต้นสี่หมู่ (จตุรยางศ์)

เพื่อความเข้าใจ กล่องวรรณยุกต์ในที่นี้จะมีลักษณะเป็นตาราง โดยกำหนดให้แกน X เป็นการกำกับรูปวรรณยุกต์ และแกน Y เป็นพยัญชนะต้นแบ่งตามอักษรสามหมู่ (ไตรยางศ์) ได้แก่ อักษรสูง อักษรกลาง และอักษรต่ำ ในแต่ละช่องของตารางเป็นคำที่ตรงกับเงื่อนไขของทั้งสองแกน แล้วจึงจัดหมวดหมู่ว่าช่องใดบ้างที่มีเสียงวรรณยุกต์เดียวกัน ตัวอย่างด้านล่างใช้สีเพื่อแสดงเสียงวรรณยุกต์ โดยสีเดียวกันแสดงเสียงวรรณยุกต์เดียวกัน ผลลัพธ์ของกล่องวรรณยุกต์ คือ เงื่อนไขการเกิดแต่ละวรรณยุกต์และจำนวนหน่วยเสียงวรรณยุกต์

ยกตัวอย่างกล่องวรรณยุกต์ของภาษาไทยกลางในตารางที่ 1

**ตารางที่ 1**. กล่องวรรณยุกต์คำไทย

|  | ไม่มีรูปวรรณยุกต์ | รูปเอก | รูปโท | คำตายสระเสียงยาว | คำตายสระเสียงสั้น |
|---|---|---|---|---|---|
| สูง | ขา ผม หนา | ข่า ไผ่ หมู่ | ข้าว ห้า หน้า | ขาด หาบ หมาก | ผัก หัก หลัก |
| กลาง | ตา กอ ปลา | ป่า แก่ เต่า | ป้า ตู้ กล้า | กราบ ปาก ตาก | ปิด กัด ตับ |
|  | บาน อาน ดาว | อ่าน ด่า บ่า | อ้าย บ้าน ด้าม | บีบ ดาบ อาบ | อุด เด็ก บิด |
| ต่ำ | นา มือ งู | โง่ ค่า พ่อ | ง้อ ค้า ม้า | งาบ มีด เชือก | งก ลัก มด |

เนื่องจากช่องที่มีสีเดียวกันคือมีเสียงวรรณยุกต์เดียวกัน จึงสามารถระบุได้ว่าภาษาไทยกลางมีทั้งหมด 5 เสียงวรรณยุกต์ พร้อมกับรูปแบบการสะกดว่าเสียงวรรณยุกต์แต่ละเสียงเกิดกับการสะกดแบบใด ระบบภาษาไทยกลางตั้งชื่อทั้ง 5 วรรณยุกต์นี้ว่า เสียงสามัญ (ช่องสีเขียว) เสียงเอก (ช่องสีเทา) เสียงโท (ช่องสีม่วงหม่น) เสียงตรี (ช่องสีม่วงสว่าง) และเสียงจัตวา (ช่องสีดำ)

กล่องวรรณยุกต์ของ Gedney (1972) สามารถนำมาใช้เพื่อทำความเข้าใจวรรณยุกต์ในภาษาอีสานได้ เนื่องจากเสียงวรรณยุกต์ในภาษาอีสานมีพยัญชนะต้นกับรูปวรรณยุกต์เป็นเงื่อนไขเช่นเดียวกับภาษาตระกูลไทอื่น ๆ งานที่ศึกษาเสียงวรรณยุกต์ภาษาอีสานในอดีตได้ใช้กล่องวรรณยุกต์เป็นส่วนหนึ่งของระเบียบวิธีวิจัยแล้วด้วยเช่นกัน (ธนานันท์ ตรงดี, 2557; พรสวรรค์ นามวัง, 2544; พร้อมสิริ นามมุงคุณ และ ศุภกิต บัวขาว, 2563; พิณรัตน์ อัครวัฒนากุล, 2546)

เมื่อใช้กับภาษาอีสาน 6 วรรณยุกต์จะได้ผลลัพธ์ตามตารางด้านล่าง โดยมีข้อกำหนดเพิ่มเติมคือ

1. คำตายจะไม่แบ่งย่อยลงเป็นรูปวรรณยุกต์อีก เพื่อให้เป็นไปตามอักขรวิธีของภาษาไทยที่ไม่อนุญาตให้ใส่ไม้เอกกับคำตายใด ๆ
2. เนื่องจากวรรณยุกต์ภาษาอีสานไม่มีชื่อบัญญัติไว้ จึงตั้งชื่อวรรณยุกต์เป็น T1-6 โดยเรียงลำดับจากช่องซ้ายมือบนสุด และไล่ลงมาด้านล่างก่อนไปทางด้านขวา

**ตารางที่ 2**. กล่องวรรณยุกต์คำไทยในสำเนียงอีสาน 6 วรรณยุกต์

|  | ไม่มีรูปวรรณยุกต์ | รูปเอก | รูปโท | คำตายสระเสียงยาว | คำตายสระเสียงสั้น |
|---|---|---|---|---|---|
| สูง | ขา ผม หนา [T1] | ข่า ไผ่ หมู่ [T4] | ข้าว ห้า หน้า [T5] | โขด หีบ หมาก [T5] | ผัก หัก หลัก [T1] |
| กลาง | ตา กอ ปลา [T2] | ป่า แก่ เต่า [T4] | ป้า ตู้ กล้า [T6] | กราบ ปูด ตอก [T5] | ปิด กัด ตับ [T1] |
|  | บาน อาน ดาว [T2] | อ่าน ด่า บ่า [T4] | อ้าย บ้าน ด้าม [T6] | บีบ ดาบ แอบ [T5] | อุด เด็ก บิด [T1] |
| ต่ำ | นา มือ งู [T3] | โง่ ค่า พ่อ [T4] | ง้อ ค้า ม้า [T6] | งาบ มีด เชือก [T6] | งก ลัก มด [T4] |

ดังนั้นจึงสามารถสรุปได้ว่าเสียงในภาษาอีสาน 6 วรรณยุกต์ มีปัจจัยที่กำหนดเสียงวรรณยุกต์ คือ

- รูปวรรณยุกต์
- หมู่ของพยัญชนะต้น
- การเป็นคำเป็น คำตายสระเสียงสั้น หรือคำตายสระเสียงยาว

วรรณยุกต์ของภาษาอีสาน 6 วรรณยุกต์นี้มีเงื่อนไขการเกิดแตกต่างกับภาษาไทยกลางอย่างชัดเจน (ธนานันท์ ตรงดี, 2557) ยกตัวอย่างเช่น คำใด ๆ ที่มีรูปวรรณยุกต์เอกในภาษาอีสานจะมีเสียงวรรณยุกต์เดียวกัน ในขณะที่คำที่มีรูปวรรณยุกต์เอกในภาษาไทยกลางจะมีเสียงวรรณยุกต์ต่างกันไปตามพยัญชนะต้น (อักษรต่ำรูปวรรณยุกต์เอกเป็นเสียงโท ในขณะที่อักษรอื่น ๆ เป็นเสียงเอก)

ในการจัดทำชุดข้อมูล หากไม่มีการกำหนดหลักเกณฑ์ที่ชัดเจนในการเลือกรูปวรรณยุกต์ ชุดข้อมูลจะเกิดความลักลั่น เนื่องจากผู้ใช้ภาษาอาจสะกดคำโดยใช้หลักการสะกดตามระบบไทยกลางหรือระบบอีสานก็ได้ ดังนั้น การจัดการกับระบบวรรณยุกต์ที่ต่างกันจึงจำเป็นต้องมีหลักการสะกดอย่างชัดเจน ดังที่จะอธิบายต่อไปในหัวข้อที่ 4.2

## 3.3 การปฏิภาค

การปฏิภาค (Correspondence) หมายถึง การเปลี่ยนเสียงหรือการกลายเสียงอย่างเป็นระบบระหว่างภาษาหนึ่งกับอีกภาษาหนึ่ง ในที่นี้หมายถึง การเปลี่ยนเสียงหรือการกลายเสียงอย่างเป็นระบบระหว่างภาษาอีสานกับภาษาไทยกลาง (วิไลศักดิ์ กิ่งคำ, 2550) ซึ่งจะทราบได้ก็ต่อเมื่อนำทั้งสองภาษามาเทียบกัน การตรงกันของเสียงพยัญชนะจะเรียกว่า เสียงปฏิภาคพยัญชนะ (consonant correspondence) และการตรงกันของเสียงสระเรียกว่า เสียงปฏิภาคสระ (vowel correspondence)

ตารางที่ 3 ยกตัวอย่างการปฏิภาคเสียงพยัญชนะระหว่างภาษาไทยกลางและภาษาอีสาน ในตัวอย่างนี้ เสียงพยัญชนะต้น ร ในภาษาไทยกลาง จะเป็นเสียงพยัญชนะต้น ฮ ในภาษาอีสาน การเปลี่ยนเสียงพยัญชนะนี้เกิดขึ้นอย่างเป็นระบบจึงพิจารณาเป็นเสียงปฏิภาคพยัญชนะ

**ตารางที่ 3**. ตัวอย่างเสียงปฏิภาคพยัญชนะระหว่างภาษาไทยกลางและภาษาอีสาน

| การออกเสียงในภาษาไทยกลาง | การออกเสียงในภาษาอีสาน |
|---|---|
| รัก | ฮัก |
| รู้ | ฮู้ |
| เรา | เฮา |

ตารางที่ 4 ยกตัวอย่างตัวอย่างเสียงปฏิภาคสระระหว่างภาษาไทยกลางและภาษาอีสาน ในตัวอย่างนี้ เสียงสระ อี ในภาษาไทยกลาง จะเป็นเสียงสระ เออ ในภาษาอีสาน การเปลี่ยนเสียงสระนี้เกิดขึ้นอย่างเป็นระบบจึงพิจารณาเป็นเสียงปฏิภาคสระ

**ตารางที่ 4.** ตัวอย่างเสียงปฏิภาคสระระหว่างภาษาไทยกลางและภาษาอีสาน

| การออกเสียงในภาษาไทยกลาง | การออกเสียงในภาษาอีสาน |
|---|---|
| ผึ้ง | เผิ้ง |
| ลึก | เลิก |
| ดึก | เดิก |

อย่างไรก็ตาม การปฏิภาคนี้นำไปสู่สองประเด็นที่ต้องพิจารณาเมื่อจัดทำชุดข้อมูล คือ

**ประเด็นที่หนึ่ง** สำหรับการปฏิภาคบางกรณี หากสะกดตามเสียงที่ผู้พูดออกเสียงจริง จะได้รูปสะกดที่ไม่คุ้นชินสำหรับผู้ใช้ภาษา เช่น "สุกเสิน" ในภาษาอีสานเป็นเสียงปฏิภาคของคำว่า "ฉุกเฉิน" ในภาษาไทยกลาง (เปลี่ยนเสียง ฉ เป็นเสียง ส)

**ประเด็นที่สอง** สำหรับการปฏิภาคบางกรณี หากสะกดตามเสียงที่ผู้พูดออกเสียงจริง จะได้รูปสะกดที่ทับซ้อนกับคำที่มีความหมายอยู่แล้วในภาษาไทยกลาง เช่น "เสีย" ในภาษาอีสานเป็นเสียงปฏิภาคของคำว่า "เสือ" ในภาษาไทยกลาง (เปลี่ยนเสียง เอือ เป็นเสียง เอีย) ซึ่งรูปเขียนซ้ำกับคำว่า 'เสีย' ที่หมายถึงสูญไป สิ้นไป

ด้วยเหตุนี้ การจัดทำชุดข้อมูลจึงต้องหาหลักการการสะกดคำที่มีการปฏิภาคที่ผู้ใช้ภาษาอีสานคุ้นชินและในขณะเดียวกันก็มีความกำกวมน้อยที่สุด

# 4 การจัดทำชุดข้อมูล

## 4.1 การเก็บข้อมูลเสียง

เพื่อให้ได้ข้อมูลเสียงที่ไม่ใช่เสียงอ่าน มีปรากฏการณ์ทางเสียงที่เกิดขึ้นเมื่อผู้พูดพูดแบบสนทนา และมีการปรากฏคำศัพท์เฉพาะถิ่นตามบริบทตามการใช้จริงของเจ้าของภาษา การเก็บข้อมูลจึงใช้วิธีการสร้างสิ่งเร้า (stimuli) ที่มุ่งเน้นให้ผู้บอกภาษาตอบตามความรู้สึกนึกคิดของตน (intuition) และไม่ได้มีการชี้นำเหมือนกับการให้อ่านข้อความ

เพื่อให้บรรลุวัตถุประสงค์การสร้างคำถามจึงคำนึงถึง 2 หลักการ ดังนี้

1. **ความเป็นธรรมชาติ** คำถามต้องมีความเป็นธรรมชาติในภาษาอีสาน ไม่ชี้นำให้ผู้บอกภาษาคิดเป็นภาษาไทยกลาง เนื่องด้วยผู้พูดภาษาอีสานมักใช้ภาษาไทยได้ในชีวิตประจำวันด้วย (McCargo & Hongladarom, 2004) สิ่งเร้าต้องไม่มีอิทธิพลจากภาษาไทยกลางหรือมีน้อยที่สุด และกระตุ้นให้ผู้บอกภาษาพูดภาษาอีสานอย่างเป็นธรรมชาติที่สุด

2. **ความหลากหลาย** คำถามต้องมีคำตอบได้หลากหลาย เพื่อทำให้ได้ข้อมูลที่ครอบคลุมและมีเนื้อหาซ้ำกันน้อยที่สุด สิ่งเร้าต้องเปิดโอกาสให้ผู้บอกภาษาให้คำตอบได้กว้างขวางตามประสบการณ์ ความคิด และบริบทของตนเอง

สองหลักการข้างต้นจึงทำให้เกิดวิธีการออกแบบสิ่งเร้า ดังนี้

**หนึ่ง การออกแบบสิ่งเร้าให้มีประเภทหลากหลาย** สิ่งเร้าต้องมีหลายรูปแบบ เช่น ตอบคำถามที่เป็นข้อความ อธิบายคำสำคัญ (keyword) ตอบคำถามจากเสียงที่ได้ยิน อธิบายหรือตอบคำถามรูปภาพ/วิดีโอ เติมบทสนทนาในสถานการณ์สมมุติ เป็นต้น จากนั้น จึงเลือกประเภทของสิ่งเร้าให้เหมาะกับหัวเรื่องที่ต้องการ ยกตัวอย่างเช่นภาพต่อไปนี้

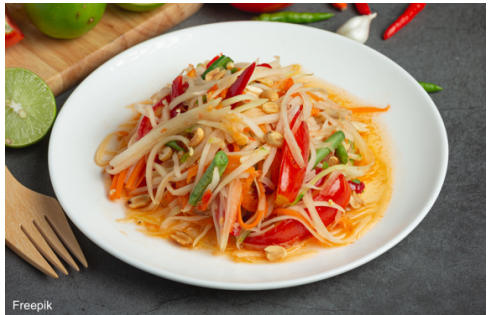

**ภาพที่ 2.** อาหารที่มีชื่อว่า "ส้มตำ" ในภาษาไทยกลาง

ในภาษาอีสาน อาหารจานนี้เรียกว่า "ตำบักหุ่ง" อย่างไรก็ตาม คนอีสานส่วนมากคุ้นชินกับคำว่า "ส้มตำ" ด้วยเช่นกันจากอิทธิพลของภาษาไทยกลาง ทั้งคนอีสานบางส่วนก็เรียกอาหารชนิดนี้ว่า "ส้มตำ" ในชีวิตประจำวัน หากถามผู้บอกภาษาอีสานว่าภาษาอีสานเรียก 'ส้มตำ' ว่าอะไรโดยตรง ผู้บอกภาษาอีสานส่วนมากอาจไม่ทันนึกถึง "ตำบักหุ่ง" เนื่องจาก "ส้มตำ" ก็เป็นคำที่คุ้นชินอยู่แล้ว เพราะฉะนั้น หากต้องการเก็บข้อมูลให้ได้ชื่ออาหารชนิดนี้ที่ตรงกับบริบทที่ใช้จริงเมื่อพูดภาษาอีสาน การเลือกใช้สิ่งเร้าเป็นรูปภาพจึงเหมาะสมที่สุดเนื่องจากไม่ชี้นำและไม่กระตุ้นให้นึกถึงภาษาไทยกลางก่อนภาษาอีสาน

**สอง การออกแบบกระบวนการเพื่อชี้นำให้ผู้บอกภาษาคิดเป็นภาษาอีสาน** ทั้งสิ่งเร้าและคำสั่งต้องชี้นำให้ผู้บอกภาษาคิดเป็นภาษาอีสาน โดยใช้การสะกดที่ทำให้ต้องอ่านข้อความเป็นสำเนียงอีสาน ไม่ใช่ภาษาไทยกลาง ยกตัวอย่างเช่น สิ่งเร้าหมวดคำถามที่เป็นข้อความสะกด

'ซุมื้อนี้เจ้าใช้บริการธนาคารอิหยังอยู่ เป็นหยังเถิงเลือกธนาคารนี้'
(คำแปล: ทุกวันนี้คุณใช้บริการธนาคารอะไรอยู่ เพราะอะไรถึงเลือกธนาคารนี้)

สังเกตว่า คำถามเลือกใช้
- 'ซุมื้อ' แทน 'ทุกวัน'
- 'เจ้า' แทน 'คุณ'
- 'เป็นหยัง' แทน 'เพราะอะไร'
- ' เถิง' แทน 'ถึง'

เพื่อให้ใกล้เคียงกับเสียงที่ผู้ใช้ภาษาอีสานใช้ในชีวิตประวัน เป็นการชี้นำให้ผู้บอกภาษาอ่านเป็นภาษาอีสานและคิดก่อนตอบเป็นภาษาอีสาน

การเก็บข้อมูลบันทึกเพศของผู้บอกภาษา ภูมิลำเนาของผู้บอกภาษา และคำถามที่ผู้บอกภาษาได้รับไว้ด้วยกัน ข้อมูลเสียงทั้งหมดได้รับการตรวจสอบว่าเป็นสำเนียงอีสานแบบ 6 วรรณยุกต์จริงและไม่มีเสียงรบกวนที่กระทบต่อความเข้าใจ เสียงที่ผ่านกระบวนการตรวจสอบทั้งหมดนำไปถอดความต่อตามหลักการที่จะอธิบายต่อไป

## 4.2   แนวทางการสะกดคำ

ปัจจุบันภาษาอีสานยังไม่มีการกำหนดหลักการสะกดอย่างเป็นทางการจากหน่วยงานใด ผู้ใช้ภาษาอีสานจึงสะกดภาษาอีสานในชีวิตประจำวันตามความรู้สึกนึกคิดของตนเองโดยไม่ได้อิงระบบการสะกดหรือการเขียนใดเป็นหลักอย่างแน่ชัด ในขณะเดียวกันผู้ใช้ภาษาอีสานก็ได้รับอิทธิพลจากภาษาไทยกลางจากการศึกษาภาคบังคับ รวมถึงอิทธิพลจากสถานะทางสังคมและเศรษฐกิจที่มาพร้อมกับใช้ภาษาไทยกลาง (McCargo & Hongladarom, 2004) การสะกดภาษาอีสานจึงมีความไม่สม่ำเสมอสูงแม้ในกลุ่มผู้ใช้ภาษาอีสานด้วย ยกตัวอย่างเช่น คำว่า "พูด" ในภาษาอีสาน ปรากฏการสะกดทั้ง 'เว่า' 'เว้า' และ 'เหว้า' ในสื่อออนไลน์

แนวทางการสะกดคำนี้จึงสร้างขึ้นโดยมีเป้าหมายเพื่อให้การสะกดคำในชุดข้อมูลภาษาอีสานมีความสม่ำเสมอและในขณะเดียวกันก็ยังคงคุ้นชินตาสำหรับผู้พูดภาษาอีสานด้วยเช่นกัน

**4.2.1 หลักการโดยรวม**

แนวทางการสะกดคำสามารถสรุปเป็นแผนผังได้ดังนี้

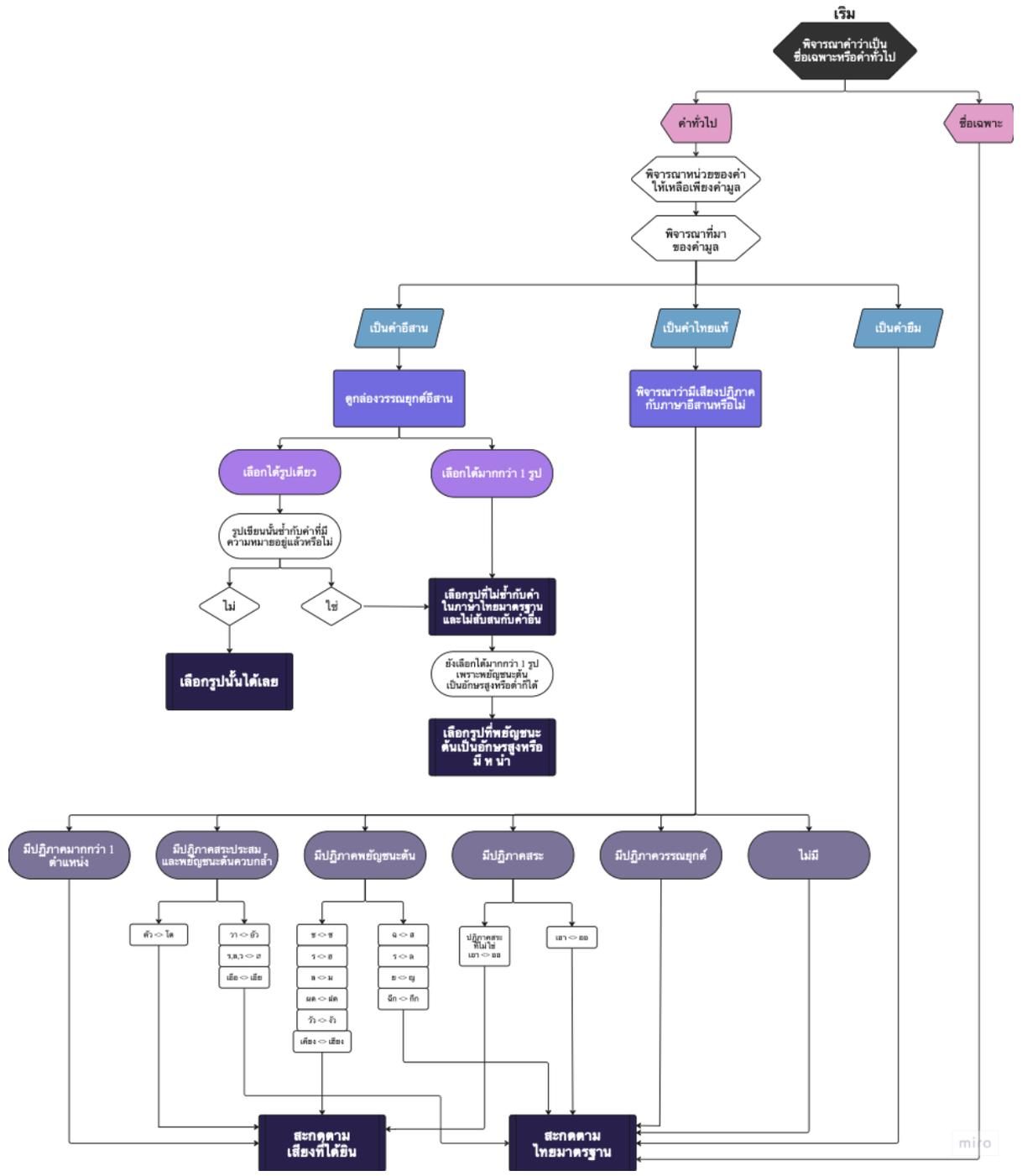

**ภาพที่ 3**. แผนผังแสดงแนวทางการสะกดคำ

โดยสรุป หลักการสะกดมีขั้นตอนดังนี้

**ขั้นตอนที่หนึ่ง** แนวทางนี้ไม่ใช้กับชื่อเฉพาะ ชื่อเฉพาะทั้งหมดคงไว้เหมือนตามที่ภาษาไทยกลางใช้ เนื่องจากรูปที่สะกดตามการออกเสียงมีโอกาสเกิดความสับสนสูง ยกตัวอย่างเช่น แม้ผู้พูดอีสานจะออกเสียงชื่อ 'ชาติชาย' ว่า "ซาดซาย" แต่หากสะกดเป็น 'ซาติซาย' หรือ 'ซาดซาย' ผู้ใช้ภาษาอาจสับสนและไม่ตระหนักว่าเป็นชื่อเฉพาะได้

**ขั้นตอนที่สอง** เมื่อตัดส่วนที่เป็นชื่อเฉพาะออก พิจารณาหน่วยของคำให้ย่อยลงเป็นระดับคำมูลเท่านั้น คำมูลหมายถึงเป็นคำที่สมบูรณ์ในตัวเอง ไม่สามารถแบ่งย่อยได้อีกแล้ว เช่น เวลา นาฬิกา ขะมักเขม้น เป็นต้น การพิจารณาคำมูลมีความสำคัญเนื่องจากหากไม่พิจารณาย่อยลงเป็นคำมูลแล้วใช้เกณฑ์อื่น จะยิ่งทำให้จัดกลุ่มข้อมูลยากขึ้นและทำให้หลักการสะกดที่เกิดขึ้นตามมาอาจจะมีความซับซ้อนขึ้นมากจนนำไปใช้จริงไม่ได้

**ขั้นตอนที่สาม** พิจารณาที่มาของแต่ละคำมูลข้างต้นตามที่มาของคำ จำแนกได้เป็น 3 กลุ่มใหญ่ ดังนี้
- **คำภาษาถิ่นอีสาน** มีข้อสังเกตคือเป็นคำที่ใช้ในถิ่นอีสานและไม่ปรากฏการใช้คำนี้ในถิ่นอื่น เช่น เหมบ (คำแปล: นอนคว่ำ) ซิด (คำแปล: สะบัดเบา ๆ) เอิ้น (คำแปล: เรียก) เหี้ยน (คำแปล: สั้น) เว้า (คำแปล: พูด)
- **คำไทย** มีข้อสังเกตคือมักเป็นคำที่มีตัวสะกดตรงตามมาตราหรือมีการใช้ไม้ม้วน เช่น มัก พูด กราบ ใกล้ ใต้
- **คำยืม** มีข้อสังเกตคือมักเป็นคำที่สะกดไม่ตรงตามมาตรา มีการใช้ไม้ทัณฑฆาต มีการใช้เสียงอักษร ห นำ หรือมีการสะกดเหมือนคำควบไม่แท้ เช่น กาล สามารถ กอล์ฟ ครรภ์ ถนน ผนัง ทรวดทรง จริง

**ขั้นตอนที่สี่** เลือกสะกดตามที่มาของคำ ดังจะอธิบายในหัวข้อถัดไป

### 4.2.2 กรณีที่เป็นคำยืม

กรณีที่เป็นคำยืม ให้ใช้การสะกดตามภาษาไทยกลางโดยอิงกับหลักเกณฑ์ของราชบัณฑิตยสถานทุกกรณี เนื่องจากหากสะกดตามการออกเสียงจะเกิดความไม่คุ้นชินอย่างมาก เช่น การสะกด 'รส' เป็น 'ฮส' หรือ 'ฮด' และหากพยายามสร้างกฎเกณฑ์เพื่อสะกดคำเหล่านี้ตามเสียงก็สามารถคาดเดาได้ว่าจะเป็นกฎเกณฑ์ที่มีความซับซ้อนจนไม่อาจนำไปใช้ได้จริง

ทั้งนี้ หลักการนี้ยกเว้นคำว่า "ก็" ซึ่งเป็นคำยืมภาษาเขมร เนื่องจากคุ้นชินจนเหมือนเป็นส่วนหนึ่งของภาษาอีสานไปแล้ว

### 4.2.3 กรณีที่เป็นคำไทย

คำไทยจำนวนมากจะเกิดการปฏิภาคในภาษาอีสานดังที่ได้กล่าวไปในส่วนก่อนหน้า งานนี้จึงรวบรวมรายการเสียงปฏิภาคระหว่างภาษาไทยกับภาษาอีสาน และจัดทำแบบสอบถามเพื่อตรวจสอบว่าผู้ใช้ภาษาอีสานส่วนใหญ่ยอมรับรูปสะกดตามเสียงปฏิภาคที่ออกจริงหรือรูปสะกดแบบไทยกลางมากกว่าสำหรับแต่ละรายการเสียงปฏิภาค ผลของแบบสอบถามทำให้สรุปหลักการถอดเสียงปฏิภาคได้เป็น 2 หมวด ดังนี้

**หมวดที่ 1** เสียงปฏิภาคที่ผู้ใช้ภาษาอีสานส่วนใหญ่คุ้นชินกับรูปสะกดที่สะกดตามเสียงมากกว่า แม้ว่ารูปสะกดนั้นไม่ตรงกับไทยกลาง คำที่มีเสียงปฏิภาคในหมวดนี้จะสะกดตามเสียงที่ผู้พูดออกจริง ตัวอย่างเช่น

**ตารางที่ 5.** ตัวอย่างเสียงปฏิภาคที่เลือกสะกดตามเสียงที่ผู้พูดออกจริง

| คู่เสียงปฏิภาค | การสะกดในภาษาไทยกลาง | การออกเสียงในภาษาอีสาน | การสะกดที่กำหนดสำหรับภาษาอีสาน |
|---|---|---|---|
| เสียง ช <> เสียง ซ | ช้าง | "ซ้าง" | ซ้าง |
|  | ชุด | "ซุด" | ซุด |
| เสียง ร <> เสียง ฮ | รัก | "ฮัก" | ฮัก |
|  | โรง | "โฮง" | โฮง |
| เสียง อี <> เสียง เออะ | ลึก | "เลิก" | เลิก |
|  | ดึก | "เดิก" | เดิก |
| เสียง อา <> เสียง อัว | ข้าม | "ข้วม" | ข้วม |
|  | สาย | "สวย" | สวย |

**หมวดที่ 2** เสียงปฏิภาคที่ผู้ใช้ภาษาอีสานส่วนใหญ่คุ้นชินกับรูปสะกดตามไทยกลางมากกว่า แม้ว่ารูปสะกดนั้นไม่ตรงกับเสียงที่ออก คำที่มีเสียงปฏิภาคในหมวดนี้จะสะกดตามไทยกลาง ตัวอย่างเช่น

**ตารางที่ 6.** ตัวอย่างเสียงปฏิภาคที่เลือกสะกดตามไทยกลาง

| คู่เสียงปฏิภาค | การสะกดในภาษาไทยกลาง | การออกเสียงในภาษาอีสาน | การสะกดที่กำหนดสำหรับภาษาอีสาน |
|---|---|---|---|
| เสียง ฉ <> เสียง ส | ฉีก | "สีก" | ฉีก |
|  | เฉย | "เสย" | เฉย |
| เสียง ร <> เสียง ล | รัก | "ลัก" | รัก |
|  | รวย | "ลวย" | รวย |
| เสียง เอือ <> เสียง เอีย | เบื่อ | "เบี่ย" | เบื่อ |
|  | เดือด | "เดียด" | เดือด |
| เสียงควบกล้ำ ว <> เสียง อัว | กวาด | "กวด" | กวาด |
|  | ความ | "ควม" | ความ |

ทั้งนี้ งานนี้ตัดประเด็นเสียงวรรณยุกต์ที่ปฏิภาคออกไป (เช่น กบ-ก๊บ) เพราะเห็นว่าน่าจะคุ้นชินกับรูปไทยกลางอยู่แล้ว การมีวรรณยุกต์กำกับที่ต่างไปจากที่คุ้นชินจะทำให้การสะกดดูซับซ้อนมากขึ้น

### 4.2.4 กรณีที่เป็นคำถิ่นอีสาน

### 4.2.4.1 การใช้กล่องวรรณยุกต์

คำถิ่นอีสานที่พจนานุกรมฉบับราชบัณฑิตสถานหากมีบัญญัติไว้แล้วจะสะกดตามพจนานุกรม เช่น "แซ่บ" (อร่อย) "อีหลี" (จริงๆ) "คิง" (ร่างกาย) เนื่องจากรูปสะกดเหล่านี้มีความคุ้นชินแล้วสำหรับผู้ใช้ภาษาอีสาน ส่วนคำที่ไม่ถูกบัญญัติไว้นั้นจะใช้ "กล่องวรรณยุกต์" (Gedney, 1972) เพื่อเป็นการสร้างหลักการสะกดที่เป็นระบบดังที่อธิบายไปแล้วในส่วนก่อนหน้า

ตารางที่ 7. กล่องวรรณยุกต์คำไทยในสำเนียงอีสาน

|  | ไม่มีรูปวรรณยุกต์ | รูปเอก | รูปโท | คำตายสระเสียงยาว | คำตายสระเสียงสั้น |
|---|---|---|---|---|---|
| สูง | ขา ผม หนา [T1] | ข่า ไผ่ หมู่ [T4] | ข้าว ห้า หน้า [T5] | โขด หีบ หมาก [T5] | ผัก หัก หลัก [T1] |
| กลาง | ตา กอ ปลา [T2] | ป่า แก่ เต่า [T4] | ป้า ตู้ กล้า [T6] | กราบ ปูด ตอก [T5] | ปิด กัด ตับ [T1] |
|  | บาน อาน ดาว [T2] | อ่าน ด่า บ่า [T4] | อ้าย บ้าน ด้าม [T6] | บีบ ดาบ แอบ [T5] | อุด เด็ก บิด [T1] |
| ต่ำ | นา มือ งู [T3] | โง่ ค่า พ่อ [T4] | ง้อ ค้า ม้า [T6] | งาบ มีด เชือก [T6] | งก ลัก มด [T4] |

ในการใช้กล่องวรรณยุกต์เพื่อสะกดคำอีสาน ผู้พูดอีสานจะออกเสียงคำที่ต้องการสะกด แล้วจึงเทียบเคียงเสียงวรรณยุกต์ในคำคำนั้นกับเสียงวรรณยุกต์ของคำในละกล่องเพื่อตรวจสอบว่าเสียงวรรณยุกต์คำที่ต้องการสะกดนั้นตรงกับการสะกดด้วยพยัญชนะต้นหมวดใดและรูปวรรณยุกต์ใด

ตัวอย่างการใช้กล่องวรรณยุกต์เพื่อเลือกสะกดคำ เช่น คำเฉพาะถิ่นอีสานที่มีความหมายว่า "ฉัน"

- ถ้าเป็น T1 จะสะกดได้ว่า 'ขอย' (พยัญชนะต้นอักษรสูง ไม่มีรูปวรรณยุกต์)
- เป็น T2 ไม่ได้ เพราะเสียง ข/ค ไม่สามารถใช้อักษรกลางตัวใดแทนได้เลย
- ถ้าเป็น T3 จะสะกดได้ว่า 'คอย' (พยัญชนะต้นอักษรต่ำ ไม่มีรูปวรรณยุกต์)
- ถ้าเป็น T4 จะสะกดได้ว่า 'ข่อย' หรือ 'ค่อย' (พยัญชนะต้นอักษรสูงหรือต่ำ รูปวรรณยุกต์เอก)
- ถ้าเป็น T5 จะสะกดได้ว่า 'ข้อย' (พยัญชนะต้นอักษรสูง รูปวรรณยุกต์โท)
- ถ้าเป็น T6 จะสะกดได้ว่า 'ค้อย' (พยัญชนะต้นอักษรต่ำ รูปวรรณยุกต์โท)

โดยเมื่อเทียบเสียงแล้ว พบว่าคำที่มีความหมายว่า "ฉัน" มีเสียงวรรณยุกต์ตรงกับ T5 จึงสะกดด้วยพยัญชนะต้นอักษรสูง รูปวรรณยุกต์โท ได้เป็น 'ข้อย' เท่านั้น

ในกรณีที่กล่องวรรณยุกต์สามารถสะกดได้หลายรูป มีข้อกำหนดการสะกดคำเพิ่มเติม ดังนี้

1. หากยังคงใช้ได้หลายรูปและไม่มีรูปซ้ำกับไทยกลาง ให้พิจารณาย้อนไปถึงคำต้นทางและเลือกรูปที่ไม่ทำให้เข้าใจผิดว่ามีความหมายอื่น เช่น "หม่า" หรือ "ม่า" ที่แปลว่า "หมัก" ให้เลือกสะกด 'หม่า' เพราะ 'หม่า' มีรูปเขียนใกล้เคียงกับ 'หมัก'
2. ในกรณีที่ย้อนไปถึงคำต้นทางไม่ได้ ให้เลือกรูปอักษรสูงหรือใช้ ห นำ เช่น "ส่ำ" หรือ "ซ่ำ" ที่แปลว่า "เท่า" ให้เลือก 'ส่ำ'

#### 4.2.4.2 การเลือกสะกดในกรณีกำกวม

1. หากสามารถใช้สระได้มากกว่า 1 เสียงโดยไม่ทำให้ความหมายแตกต่าง เช่น "เซื่อง" และ "เชี่ยง" ที่หมายถึง "ซ่อน" และทั้ง 2 รูปนี้ไม่มีใช้ในภาษาไทยกลาง จะสะกดตามการออกเสียงจริง
2. หากสามารถใช้ได้ทั้งรูปสระสั้นและยาว จะพิจารณาเลือกรูปที่สามารถสืบย้อนความหมายได้ เช่น "ซุมื้อ" กับ "ซูมื้อ" ที่แปลว่า "ทุกวัน" เพื่อให้สืบไปหาคำว่า 'ทุก' ได้ จึงควรเป็นรูปเสียงสั้นและเลือกสะกด 'ซุมื้อ'
3. หากสามารถออกเสียงเป็นได้ทั้งรูปสระสั้นยาวและสืบย้อนความหมายไม่ได้ จะเลือกรูปที่ไม่ขัดกับอักขรวิธีการสะกดคำในภาษาไทยกลาง เช่น คำว่า "ข้วน" แม้จะออกเสียงสั้นเหมือนสระอัวะ เลือกสะกด 'ข้วน' จะไม่ใช้รูป 'ขั้วน'

หลักเกณฑ์การสะกดคำที่กำหนดขึ้นนี้ มิได้เป็นการสร้างอักขรวิธีใหม่ ซึ่งหลักเกณฑ์นี้กำหนดให้ไม่ขัดกับอักขรวิธีการเขียนสะกดคำในภาษาไทยกลาง เช่น

- ไม่ใช้ไม้ไต่คู้พร้อมกับเครื่องหมายวรรณยุกต์
- สระลดรูป เช่น สระ อัว เมื่อมีตัวสะกดจะไม่ใส่ไม้หันอากาศ เช่น 'ข้วน' (ไม่สะกดว่า 'ขั้วน')
- ไม่ใช้ไม้ตรีและไม้จัตวากับอักษรสูงและอักษรต่ำ เพื่อให้สอดคล้องกับการใช้กล่องวรรณยุกต์ โดยจะสงวนไว้ใช้กับอักษรกลางเท่านั้น เช่น 'ดู๋' (คำถิ่นอีสาน แปลว่า ขยัน, มาก, บ่อย) นอกจากนี้ คำที่ใช้ไม้ตรีและไม้จัตวาส่วนใหญ่เป็นคำยืมหรือคำเลียนเสียงธรรมชาติ เช่น ก๋วยเตี๋ยว โป๊ ป๊าด

## 4.3 แนวทางการถอดความ

การถอดความตามสิ่งที่พูดในสถานการณ์การพูดจริงมักจะมีความไม่สม่ำเสมอสูง เนื่องจากคำพูดคำพูดหนึ่งมักจะสามารถถอดความได้หลายแบบโดยยังคงให้ความหมายเดิม ยกตัวอย่างเช่น คำว่า 'ครับ' สามารถมีการขึ้นลงของน้ำเสียง การแสดงความลังเล หรือการกร่อนเสียง กลายเป็นเสียง "คับ" "ขับ" "ค้าบ" หรือ "คร้าบ" ก็ได้โดยยังคงความหมายเดิม อีกหนึ่งตัวอย่างคือคำพูด "คิวอาร์โค้ด" สามารถถอดความเป็น 'คิวอาร์โค้ด' 'คิวอาร์โคด' 'QR Code' หรือ 'QR code' ก็ได้โดยยังคงถูกต้องตามเสียงอยู่

แนวทางการถอดความนี้จึงกำหนดขึ้นเพื่อแก้ปัญหาดังกล่าวที่มักเกิดขึ้นเมื่อมีผู้ถอดความมากกว่าหนึ่งคน ซึ่งจะช่วยให้ชุดข้อมูลที่ออกมามีความสม่ำเสมอสูงสุด

แนวทางการถอดความสามารถจัดกลุ่มเป็น แนวทางการถอดความทั่วไป และ แนวทางการถอดความที่ว่าด้วยปรากฏการณ์ทางเสียง ในที่นี้จะเปรียบเทียบกับปรากฏการณ์ในภาษาไทยกลางเมื่อมีตัวอย่างภาษาอีสานเพื่อให้เข้าใจง่ายขึ้นสำหรับผู้ที่ไม่พูดภาษาอีสาน

## 4.3.1 แนวทางการถอดความโดยทั่วไป

**การสะกด** สะกดตามแนวทางการสะกดที่ได้อธิบายไปในหัวข้อ 4.2

**เครื่องไม้วรรคตอน ไม้ยมก ตัวเลข** ไม่ใช้เครื่องหมายวรรคตอน ไม้ยมก หรือตัวเลข (ยกเว้นจุดในตัวย่อและกรณีที่เป็นชื่อเฉพาะ) เพื่อลดความกำกวมและความไม่สม่ำเสมอ เช่น ถอดเป็น '*สส. จำนวนสองร้อยคนเดินเข้าเข้าออกออกร้าน 7-Eleven*'

**การเว้นวรรค** เว้นวรรคตามความเหมาะสมโดยอิงตามหลักราชบัณฑิตยสถาน ยกเว้นเรื่องเว้นวรรคใหญ่กับเว้นวรรคเล็ก ในที่นี้ให้ใช้การเว้นวรรค 1 เคาะเสมอ โดยทั่วไปเว้นวรรคเพื่อแบ่งวลีหรือประโยค เว้นหน้าและหลังตัวย่อ

หากเป็นตัวเลขหลายตัวต่อกัน ให้เว้นวรรคตามที่ผู้พูดหยุดจริง เช่น ผู้พูดพูด "สนใจติดต่อ[เว้น]ศูนย์แปดสี่[เว้น]สามหกสอง[เว้น]สามหกสองสี่" ถอด '*สนใจติดต่อ ศูนย์แปดสี่ สามหกสอง สามหกสองสี่*' อีกหนึ่งตัวอย่างคือหากผู้พูดพูด "โทรด่วนก่อนหมดเขต[เว้น]ศูนย์สองแปดสามหกเจ็ดเจ็ดเจ็ดเจ็ด" ถอด '*โทรด่วนก่อนหมดเขต ศูนย์สองแปดสามหกเจ็ดเจ็ดเจ็ดเจ็ด*'

**อักษรย่อ ตัวย่อ** ถอดความตามหลักราชบัณฑิตยสถานหรือการเขียนอย่างเป็นทางการของเจ้าของหน่วยงาน เช่น ผู้พูดพูด "ตอนเดือน มอคอ ผู้เข้ามาร่วมมาจากมอชอบ้าง มอศอวอบ้าง ปอปอชอบ้าง" ถอด '*ตอนเดือน ม.ค. คนที่เข้ามาร่วมมาจาก มช. บ้าง มศว บ้าง ป.ป.ช. บ้าง*'

**การเลือกทับศัพท์หรือตัวอักษรละติน** การเลือกสะกดคำภาษาต่างประเทศด้วยตัวอักษรภาษาไทยหรือภาษาอังกฤษจะพิจารณาว่าคำคำนั้นเป็นคำทั่วไปหรือคำเฉพาะ เพื่อเลือกรูปที่มีความคุ้นชินมากที่สุดและมีความกำกวมน้อยที่สุด

## 4.3.2 แนวทางการถอดความที่ว่าด้วยปรากฏการณ์ทางเสียง

**การตัดพยางค์** การตัดพยางค์ หมายถึง การไม่ออกเสียงบางพยางค์ของคำคำหนึ่ง ในกรณีนี้ จะถอดความตรงตามสิ่งที่ได้ยิน

ตารางที่ 8. ตัวอย่างการตัดพยางค์ในภาษาอีสาน

| เสียงที่ได้ยิน | คำเต็ม | ข้อความถอด |
| --- | --- | --- |
| ตำหุ่ง | ตำบักหุ่ง | ตำหุ่ง |
| หยัง | อิหยัง | หยัง |
| ยโส | ยโสธร | ยโส |

ตารางที่ 9. ตัวอย่างการตัดพยางค์ในภาษาไทยกลาง

| เสียงที่ได้ยิน | คำเต็ม | ข้อความถอด |
| --- | --- | --- |
| ไร | อะไร | ไร |
| มหาลัย | มหาวิทยาลัย | มหาลัย |

**การกลมกลืนเสียง** การกลมกลืนเสียง หมายถึง การเปลี่ยนเสียงพยัญชนะหรือสระของบางพยางค์ให้มีความใกล้เคียงกับพยางค์ก่อนหน้าหรือถัดไปมากขึ้น เพื่อให้ออกเสียงง่ายขึ้น ในกรณีนี้ การกลมกลืนเสียงโดยจำนวนพยางค์ยังคงเดิม จะถอดความตามเสียงที่ได้ยิน หากกลมกลืนเสียงโดยทำให้จำนวนพยางค์เปลี่ยนไป จะถอดความตามรูปปกติ

**ตารางที่ 10.** ตัวอย่างการกลมกลืนเสียงที่จำนวนพยางค์เท่าเดิม

| เสียงที่ได้ยิน | ข้อความถอด |
|---|---|
| ยังไง<br>(เปลี่ยนเสียง ร ใน "ไร" เป็นเสียง ง) | ยังไง |
| อย่างงั้น<br>(เปลี่ยนเสียง น ใน "นั้น" เป็นเสียง ง) | อย่างงั้น |

**ตารางที่ 11.** ตัวอย่างการกลืมกลืนเสียงที่จำนวนพยางค์เปลี่ยนไป

| เสียงที่ได้ยิน | ข้อความถอด |
|---|---|
| ยิ่บเอ็ด | ยี่สิบเอ็ด |
| เตง | ตัวเอง |

**การขึ้นเสียงสูง การลากเสียง และการกร่อนเสียง** การขึ้นเสียงสูง หมายถึง การออกเสียงคำคำหนึ่งให้มีระดับสูงกว่าปกติ การลากเสียง หมายถึง การยืดระยะเวลาของคำคำหนึ่งให้ยาวกว่าปกติ และการกร่อนเสียง หมายถึง การออกเสียงคำคำหนึ่งให้ระยะเวลาสั้นกว่าปกติ โดยการกร่อนเสียงมักจะเกิดขึ้นพร้อมกับการเปลี่ยนวรรณยุกต์ด้วย ในกรณีนี้ ให้สะกดตามเสียงเมื่อออกเสียงคำคำนั้นอย่างตั้งใจ (กล่าวคือการออกเสียงปกติที่ไม่มีการขึ้นเสียงสูง การลากเสียง และการกร่อนเสียง)

**ตารางที่ 12.** ตัวอย่างการถอดความการขึ้นเสียงสูง การลากเสียง และการกร่อนเสียงในภาษาอีสาน

| เสียงที่ได้ยิน | ข้อความถอด | คำอธิบาย |
|---|---|---|
| ว่าจั๋งใด๋ | ว่าจั่งใด | • "จั๋ง" เป็นการขึ้นเสียงสูงของ "จั่ง" ที่พิจารณาเป็นคำถิ่นอีสาน จึงถอดตามกล่องวรรณยุกต์อีสานเมื่อออกเสียงอย่างตั้งใจ<br>• "ใด๋" เป็นการขึ้นเสียงสูงของ "ใด" ที่พิจารณาเป็นคำไทยกลาง จึงถอดความรูปพจนานุกรม |
| ไวแถะ | ไวแท้ | "แถะ" เป็นการกร่อนเสียงของ "แท้" ที่พิจารณาเป็นคำไทยกลาง จึงถอดความตามรูปพจนานุกรม |
| บ่ไปแล้วบาดนี้ | บ่ไปแล้วบัดนี้ | "บาด" เป็นการลากเสียงของ "บัด" ที่พิจารณาเป็นคำไทยกลาง จึงถอดความรูปพจนานุกรม |

**ตารางที่ 13.** ตัวอย่างการถอดความการขึ้นเสียงสูง การลากเสียง และการกร่อนเสียงในภาษากลาง

| เสียงที่ได้ยิน | ข้อความถอด | คำอธิบาย |
|---|---|---|
| บ๊า ไม่มีอะไร | บ้า ไม่มีอะไร | "บ๊า" เป็นการขึ้นเสียงสูงของ "บ้า" |
| ใช่ป่าว | ใช่เปล่า | "ป่าว" เป็นการกร่อนเสียงของ "เปล่า" |
| ได้ค้าบ | ได้ครับ | "ค้าบ" เป็นการลากเสียงของ "ครับ" |

จากหลักการข้างต้น ทำให้คำดัชนีปริจเฉท (discourse markers) มีการกำหนดรูปเขียนสำหรับคำหมวดต่อไปนี้

**คำลงท้ายหรือคำอนุภาค**

**ตารางที่ 14.** ตัวอย่างการถอดความลงท้ายหรือคำอนุภาคในภาษาอีสาน

| การออกเสียงที่เป็นไปได้ | รูปเขียนที่กำหนด | คำแปลในภาษาไทยกลาง |
|---|---|---|
| ไป่ ไป้ ไป๊ ไป๋ | ไป่ | หรือยัง |
| เบาะ บ่อ บ้อ บ๊อ | บ่ | ใช่ไหม |
| ดุ ดุ๋ ดุ๊ ดู ดู๋ ดู๊ ดู | ดุ | สิ |
| เด้อ เด๋อ เด๋อะ เด๊อะ | เด้อ | นะ |
| ตั๋ว ตั้ว ตั๊ว ตั๊วะ ตั๋วะ ตั้วะ ตั๊วะ | ตั๋ว | นะ |

**คำเติมเต็ม คำคั่นบทสนทนา หรือคำอุทาน**

**ตารางที่ 15.** ตัวอย่างการถอดความคำเติมเต็ม คำคั่นบทสนทนา หรือคำอุทานในภาษาอีสาน

| การออกเสียงที่เป็นไปได้ | รูปเขียนที่กำหนด | คำแปลในภาษาไทยกลาง |
|---|---|---|
| ห่วย ฮ่วย ฮ้วย | ฮ่วย | เฮ้ย |
| โอ๋ติ โอ๊ติ โอ๋ติ๋ โอ๋ติ้ | โอ๋ติ | เอ้าเหรอ |
| ป๊าด ป้าด ป่าด | ป๊าด | โอ้โห |

**คำแสดงความลังเล**

**ตารางที่ 16.** ตัวอย่างการถอดความคำแสดงความลังเลในภาษาอีสาน

| การออกเสียงที่เป็นไปได้ | รูปเขียนที่กำหนด | คำแปลในภาษาไทยกลาง |
|---|---|---|
| อั่น อั้น อั๊น | อั่น | เอิ่ม |
| อา อ่า อ๊า | อ่า | เออ |

## 4.4 แนวทางการถ่ายถอดเสียงสำหรับพจนานุกรมคำอ่าน

งานนี้กำหนดแนวทางการถ่ายถอดเสียงเป็นสัทอักษรไว้ด้วยเพื่อใช้เป็นแนวทางในการนำไปใช้พัฒนาต่อยอด เช่น ใช้ในกรณีที่โมเดลต้องการเรียนรู้การออกเสียงผ่านสัทอักษรเพื่อให้สามารถถอดความคำที่ไม่เคยเรียนรู้มาก่อนได้ แนวทางการถ่ายทอดเสียงนี้จะพ้องไปกับการแปลงรูปเขียนของคำเป็นรูปทางเสียง (grapheme-to-phoneme conversion) ของ PyThaiNLP (Phatthiyaphaibun et al., 2023) เนื่องจากเป็นการถ่ายถอดเสียงที่ใช้กันอย่างแพร่หลายในปัจจุบันและจึงจะทำให้เกิดการประยุกต์ใช้ได้กว้างขวางที่สุด พจนานุกรมนี้จะเริ่มจากการถ่ายถอดเสียงคำมูลก่อนเท่านั้นเพื่อเป็นพื้นฐานก่อนรวบรวมคำที่มีความซับซ้อนมากขึ้นอย่างเช่นคำประสม

ขั้นตอนการถ่ายถอดเสียงมีดังต่อไปนี้

**ขั้นตอนที่หนึ่ง กำหนดโครงสร้างพยางค์** เนื่องจากภาษาอีสานมีโครงสร้างพยางค์เหมือนภาษาไทย จึงสามารถเขียนโครงสร้างได้ดังนี้

$$C(C)V(V)(C)T$$

โดยกำหนดให้ C แทนเสียงพยัญชนะ ให้ V แทนเสียงสระ และให้ T แทนเสียงวรรณยุกต์ กล่าวคือ
- ต้องมีพยัญชนะต้นอย่างน้อย 1 หน่วยเสียงแต่ไม่เกิน 2 หน่วยเสียง
- ต้องมีสระเดี่ยวหรือสระประสม
- มีหรือไม่มีพยัญชนะท้ายก็ได้
- ต้องมีวรรณยุกต์กำกับเสมอ

โดยจุดที่แตกต่างกับภาษาไทยคือคลังหน่วยเสียงที่ไม่เหมือนกัน

**ขั้นตอนที่สอง กำหนดคลังหน่วยเสียง** โดยงานนี้เลือกกำหนดคลังหน่วยเสียง ดังนี้

**หน่วยเสียงพยัญชนะต้นเดี่ยว**

ตารางที่ 17. หน่วยเสียงพยัญชนะต้นเดี่ยวภาษาอีสาน

| หน่วยเสียง | สัทอักษร | ตัวอย่างคำ | หมายเหตุ |
|---|---|---|---|
| ป | p | ป่า, เป็ด | |
| พ/ผ | $p^h$ | ผ่า, พี (คำแปล: อ้วน) | |
| บ | b | บัง, เบ็ด | |
| ต | t | ต่อ, ตับ | |
| ท/ถ | $t^h$ | ถอย, ทอด | |
| ด | d | ดอง, ดาก (คำแปล: ก้น) | |
| จ | t͡ɕ | จาน, จืด | |

| ช/ฉ | t͡ɕʰ | โชคชัย | ใช้กับคำยืมหรือชื่อเฉพาะ |
|---|---|---|---|
| ก | k | ก้อย, กอบ | |
| ข/ค | kʰ | ขึ้น, คืน | |
| อ | ʔ | อบ, อาบ | |
| ม | m | หมา, มีด | |
| น | n | หนอน, นอก | |
| ญ | ɲ | ย่าง (คำแปล: เดิน), ญ้อน (คำแปล: เพราะ), หญิง | ไม่มีในภาษาไทยกลาง |
| ง | ŋ | หงอก, งาม | |
| ฟ/ฝ | f | ฝัด, ฟาง | |
| ซ/ส | s | เซื่อ, เสีย | |
| ฮ/ห | h | เฮา | |
| ล | l | หลาม, ลอง, รวย | |
| ว | w | หวาย, วาง | |
| ย | j | ยา, เย็น, ญาติ | |

<u>หมายเหตุ</u> การกำหนดคลังหน่วยเสียงพยัญชนะต้นเดี่ยวนี้ไม่มีหน่วยเสียง ร /r/ แต่มีหน่วยเสียง ญ /ɲ/

**หน่วยเสียงพยัญชนะควบ**

ตารางที่ 18. หน่วยเสียงพยัญชนะควบภาษาอีสาน

| หน่วยเสียง | สัทอักษร | ตัวอย่างคำ | หมายเหตุ |
|---|---|---|---|
| คว/ขว | kʰ w | เคว้ง, ขวิด | จำกัดให้ใช้กับคำไทยแท้บางคำเท่านั้น |

<u>หมายเหตุ</u> หน่วยเสียงพยัญชนะควบนี้ไม่มีการควบกล้ำหน่วยเสียง ล /l/ และ ร /r/ แต่อนุญาตให้ใช้หน่วยเสียง ว /w/ กับคำไทยแท้บางคำเท่านั้น

**หน่วยเสียงสระ**

ตารางที่ 19. หน่วยเสียงสระภาษาอีสาน

| หน่วยเสียง | สัทอักษร | ตัวอย่างคำ | หมายเหตุ |
|---|---|---|---|
| อิ | i | ยิง | |
| อี | iː | ปืน | |
| เอะ | e | เต็ง (คำแปล: ทับ) | |
| เอ | eː | เหมบ (คำแปล: หมอบ) | |
| แอะ | ɛ | แข็ง | |
| แอ | ɛː | แหมบ (คำแปล: แบน) | |

| อะ | a | นั่ง | |
|---|---|---|---|
| อา | aː | ยาม (คำแปล: เวลา) | |
| เอาะ | ɔ | หว่อง (คำแปล: หลอ) | |
| ออ | ɔː | คอง (คำแปล: รอ) | |
| โอะ | o | อ่ง (คำแปล: หยิ่ง) | |
| โอ | oː | โตน (คำแปล: กระโดดลง) | |
| อุ | u | ยุง | |
| อู | uː | สูบ | |
| อึ | ɯ | มืน (คำแปล: ดื้อด้าน) | |
| อือ | ɯː | มืน (คำแปล: ลืมตา) | |
| เออะ | ɤ | เลิก (คำแปล: ลึก) | |
| เออ | ɤː | เหิ่ม (คำแปล: ห่าม) | |
| เอีย | i a̯ | เปียก | |
| เอือ | ɯ a̯ | เดือน, เทื่อ (คำแปล: ครั้ง) | |
| อัว | u a̯ | สวน | |

**หน่วยเสียงพยัญชนะท้าย**

ตารางที่ 20. หน่วยเสียงพยัญชนะท้ายภาษาอีสาน

| หน่วยเสียง | สัทอักษร | ตัวอย่างคำ | หมายเหตุ |
|---|---|---|---|
| แม่ กบ | p̚ | ตาบ (คำแปล: ปะ) | |
| แม่ กด | t̚ | อัด (คำแปล: ปิด) | |
| แม่ กก | k̚ | เผือก | |
| แม่ กม | m | ความ | |
| แม่ กน | n | หมาน (คำแปล: มีโชค) | |
| แม่ กง | ŋ | เมือง | |
| แม่ เกอว | w | แข้ว (คำแปล: ฟัน) | |
| แม่ เกย | j | คาย (คำแปล: ระคาย, คัน) | |
| แม่ ก กา | ʔ | เตะ | ใช้กับพยางค์สระเสียงสั้นเท่านั้น |

**หน่วยเสียงวรรณยุกต์**

ภาษาอีสานสำเนียงเป้าหมายเป็นภาษาที่มีวรรณยุกต์ 6 หน่วยเสียง แต่เนื่องจากได้รับอิทธิพลของภาษาไทยกลาง จึงเพิ่มหน่วยเสียงตรีเข้าไปเป็นหน่วยเสียงภาษาต่างถิ่น ดังนี้

**ตารางที่ 21.** หน่วยเสียงวรรณยุกต์ภาษาอีสาน

| หน่วยเสียง | สัทอักษร | ตัวอย่างคำ | หมายเหตุ |
|---|---|---|---|
| T1 | ˧˩ | ผี ฝา ถุง กัด ตก จบ | |
| T2 | ˩˩ | จาน กา ปลา ปู บิน แดง ดาว | |
| T3 | ˩˧ | มือ งู นา คาว พาน ฟัน | |
| T4 | ˦ | ไก่ ต่อ ส่าย ง่าย ว่าง ท่า | |
| T5 | ˥˩ | ข้าว ห้า หน้า ขาด หาบ หมาก | |
| T6 | ˦˩ | ป้า บ้าน ม้า งาบ มีด เชือก | |
| ตรี | ˦˥ | อุ๊ย โอ๊ย ปั๊ด | สำหรับเสียงวรรณยุกต์ที่เทียบเท่าเสียงตรีในภาษาไทยกลาง |

**ขั้นตอนที่สาม กำหนดข้อบังคับทางสัทสัมผัส** เพื่อให้เกิดความสม่ำเสมอ โดยข้อกำหนดเหล่านี้อ้างอิงจากความคุ้นชินของผู้พูดอีสาน ได้แก่

- กำหนดให้ไม่มีหน่วยเสียง ร /r/ ให้ถอดเสียงหน่วยเสียงดังกล่าวเป็นหน่วยเสียง ฮ /h/ หรือ ล /l/ ตามที่ผู้พูดภาษาอีสานส่วนใหญ่ออกจริง
- ไม่มีเสียงพยัญชนะต้นควบกล้ำ เช่น
- 'ปราบ' ถอดเสียงได้เป็น /p a: p̚ ˥˩/ (ป-อา-บ-T5)
- 'คลาย' ถอดเสียงได้เป็น /kʰ a: j ˩˧/ (ค-อา-ย-T3)
- 'ความ' ถอดเสียงได้เป็น /kʰ u a̯ m ˩˧/ (ค-อัว-ม-T3)
- 'เกวียน' ถอดเสียงได้เป็น /k i a̯ n ˩˩/ (ก-เอีย-น-T2)

ทั้งนี้ ข้อบังคับนี้ยกเว้น คำควบกล้ำ ว บางคำที่ยืมเข้ามาจากภาษาไทยกลาง ยังคงเสียงควบกล้ำ ว ไว้เนื่องจากตรงกับเสียงที่ผู้พูดอีสานส่วนใหญ่ออกเสียงจริง เช่น คำว่า "ขวิด" "เคว้ง" "คว้าง" ที่ให้คงเสียงควบไว้ได้

- สระเสียงสั้นต้องมีตัวสะกดเสมอ โดยหากไม่มีรูปตัวสะกด ให้ใช้ /ʔ/ เป็นตัวสะกดในการถอดเสียง เช่น
  - 'เกาะ' ถอดเสียงได้เป็น /k ɔ ʔ ˧˩/ (ก-เอาะ-?-T1)
  - 'พุ' ถอดเสียงได้เป็น /pʰ u ʔ ˦/ (พ-อุ-?-T4)
- เสียงพยัญชนะท้ายใช้ตามหน่วยเสียงพยัญชนะท้ายที่กำหนดไว้ข้างต้นเท่านั้น หากเป็นคำยืมจากภาษาต่างประเทศ ให้ปรับเสียงพยัญชนะท้ายที่กำหนดตามที่คนอีสานส่วนใหญ่ออกเสียงจริง

**ขั้นตอนที่สี่ ถ่ายถอดเสียง** ถอดเสียงตามการออกเสียงของผู้พูดภาษาอีสานส่วนใหญ่เมื่อออกเสียงแต่ละพยางค์โดด ๆ เสียงที่ถอดออกมาต้องสอดคล้องกับคลังหน่วยเสียงภาษาอีสาน มีองค์ประกอบให้ครบโครงสร้างพยางค์ และไม่ขัดกับข้อบังคับทางสัทสัมผัสของภาษา

**ตารางที่ 22.** ตัวอย่างการถ่ายถอดเสียงภาษาอีสาน

| คำศัพท์ | การถ่ายถอดเสียง |
|---|---|
| กา | k a : ˨˩ <br> (ก-อา-T2) |
| บอล | b ɔ : n ˨˩ <br> (บ-ออ-น-T2) |
| ซอส | s ɔ : t̚ ˦˥ <br> (ซ/ส-ออ-ด-T4) |

ทั้งนี้ บางพยางค์หรือบางคำมักจะมีเสียงวรรณยุกต์แตกต่างไปจากเมื่อออกเสียงคำคำนั้นรวดเดียวหรือในบริบทที่ใหญ่กว่าคำ ในกรณีนี้ จะถ่ายถอดเสียงที่เมื่อพยางค์นั้นออกเสียงโดด ๆ เท่านั้น เนื่องจากไม่ต้องการถ่ายถอดเสียงในระดับหน่วยเสียงย่อย (allophone) เช่น

**ตารางที่ 23.** ตัวอย่างการถ่ายถอดเสียงภาษาอีสานเพิ่มเติม

| คำศัพท์ | การถ่ายถอดเสียงตามหลักการ | การถ่ายถอดเสียงที่ไม่ตรงกับหลักการ |
|---|---|---|
| สิ | s i ʔ ˦˦ <br> (ซ/ส-อิ-?-T1) | s i ˨˩ <br> (ซ/ส-อิ-T5) |
| ศาสนา | s a : t̚ ˦˥ . s a ʔ ˦˦ . n a : ˦˦ <br> (ซ/ส-อา-ด-T5 . ซ/ส-อะ-?-T1 . น-า-T1) | s a : t̚ ˦˥ . s a ˦˥ . n a : ˦˦ <br> (ซ/ส-อา-ด T5 . ซ/ส-อะ-T5 . น-า-T1) |

**ขั้นตอนที่ห้า กำกับข้อมูลเพิ่มเติม**

**ในกรณีที่มีเสียงแปร** เสียงแปรเกิดขึ้นเมื่อคำศัพท์คำหนึ่งสามารถออกเสียงได้หลายแบบโดยที่ความหมายไม่เปลี่ยนแปลง ในกรณีนี้จะมีการเลือกรูปเสียงแปรหลักและเสียงแปรรอง โดยพิจารณาเสียงที่ผู้พูดภาษาอีสานส่วนใหญ่พูดเป็นเสียงแปรหลัก และเลือกเสียงแปรอื่น ๆ เป็นเสียงแปรรอง ยกตัวอย่างเช่น

**ตารางที่ 24.** ตัวอย่างการถ่ายถอดเสียงแปรภาษาอีสาน

| คำ/พยางค์ | การถ่ายถอดเสียง | เสียงแปร |
|---|---|---|
| รัก | h a k ˦˥ <br> (ฮ/ห-อะ-ก-T4) | l a k ˦˥ <br> (ล-อะ-ก-T4) |
| เสือ | s ɯ a̠ ˦˦ <br> (ซ/ส-เอือ-T1) | s i a̠ ˦˦ <br> (ซ/ส-เสีย-T1) |
| พ่อ | pʰ ɔ: ˧˩ <br> (พ/ผ-ออ-T4) | pʰ ɔ: ˨˩ <br> (พ/ผ-ออ-T6) |

**ในกรณีที่เป็นคำพ้องรูป** คำพ้องรูปหมายถึงคือคำที่มีรูปเขียนเหมือนกัน แต่ออกเสียงต่างกันและความหมายแตกต่างกัน คำพ้องรูปแตกต่างจากคำที่มีเสียงแปร เนื่องจากคำที่มีเสียงแปรยังคงความหมายเดิม ในขณะที่คำพ้องรูปมีความหมายแตกต่างกันและบริบทการเกิดแตกต่างกันอย่างชัดเจน ในกรณีนี้ จะถอดความตามการออกเสียงและให้บริบทการใช้งานประกอบ ยกตัวอย่างเช่น

*ตารางที่ 25. ตัวอย่างการถ่ายถอดคำพ้องรูปภาษาอีสาน*

| คำ/พยางค์ | การถอดเสียง | บริบท |
|---|---|---|
| ย่าง | j a: ŋ ˨˩ <br> (ย-อา-ง-T6) | วิธีทำอาหารแบบหนึ่ง เช่น ไก่ย่าง |
| ย่าง | ɲ a: ŋ ˧˩ <br> (ญ-อา-ง-T4) | ย่างเดิน |
| รำ | h a m ˩˦ <br> (ฮ/ห-อะ-ม-T3) | ผงเยื่อหรือละอองเมล็ดข้าวสาร เช่น รำข้าว |
| รำ | l a m ˩˦ <br> (ล-อะ-ม-T3) | การแสดงที่มีการเคลื่อนไหวร่างกาย เช่น รำไทย |

# 5 ข้อจำกัดของชุดข้อมูลและข้อเสนอแนะ

## 5.1 ข้อจำกัดของการจำแนกสำเนียง

เกณฑ์วรรณยุกต์ที่ใช้งานนี้ใช้เป็นเพียงเกณฑ์หนึ่งในการจำแนกสำเนียงเท่านั้น งานในอนาคตสามารถสำรวจการจำแนกสำเนียงโดยใช้เกณฑ์ที่ต่างไปได้

## 5.2 ข้อจำกัดของแนวทางการสะกด

**กฎเกณฑ์ซับซ้อน** ผู้ใช้แนวทางการสะกดนี้ต้องจดจำกฎเกณฑ์จำนวนมากและต้องพึ่งพาความรู้ความเข้าใจทางภาษาและภาษาศาสตร์เป็นพื้นฐานด้วย ผู้ที่ไม่ได้คลุกคลีกับข้อมูลทางภาษามากอาจเกิดความสับสนในการจำแนกข้อมูล เนื่องจากการพิจารณาหลายมิติร่วมกันทำให้เกิดความเข้าใจผิดหรือวิเคราะห์ผิดแนวทางได้ง่าย

**รูปสะกดขัดกับรูปที่คุ้นชิน** ผู้ใช้แนวทางการสะกดนี้ต้องใช้เวลาเพื่อทำความคุ้นชินกับรูปเขียนบางรูปที่ผู้ใช้ภาษาอีสานอาจคุ้นชินแล้ว เช่น ในแนวทางนี้กำหนดการสะกด "ข้อย" ด้วยไม้โท แต่ผู้ใช้ภาษาอีสานโดยทั่วไปอาจชินกับการสะกด 'ข่อย' ด้วยไม้เอกในสื่อออนไลน์แล้ว

**รูปสะกดดูมีความลักลั่น** รูปสะกดบางรูปอาจดูมีความลักลั่น เช่น การพูด "ช้าง" หากหมายถึงสัตว์จะพิจารณาเป็นคำนามทั่วไปและสะกด 'ช้าง' เช่น ใน 'ข้อยเบิ่งช้าง' (คำแปล: ฉันเห็นช้าง) ในขณะที่หากหมายถึงชื่อเฉพาะ จะสะกด 'ช่าง' เช่น 'กินเบียร์ช่าง' แม้ว่าผู้พูดจะออกเสียงแบบเดียวกันชัดเจน

งานในอนาคตสามารถสำรวจเพิ่มเติมได้ว่ามีกรณีอื่นใดอีกบ้างนอกจากเสียงปฏิภาคที่รูปสะกดไม่คุ้นตาสำหรับผู้ใช้ภาษาอีสาน เช่น การสำรวจว่าการสะกดคำยืมภาษาต่างประเทศหรือชื่อเฉพาะตามไทยกลางคุ้นชินในทุกกรณีหรือไม่ เพื่อจะพิจารณาปรับแนวทางนี้ต่อไปในอนาคต

## 5.3  ข้อจำกัดของแนวทางการถอดความ

**ผู้ถอดความต้องอาศัยการตีความ** แนวทางการถอดความนี้สามารถประยุกต์ใช้ได้ยากในกรณีที่เสียงมีความกำกวมสูง เช่น แนวทางนี้กำหนดรูปคำลงท้ายเป็น 'เดะ' (สำหรับความหมาย "นะ") และ 'เด้' (สำหรับความหมาย "ไง") ทว่าข้อมูลเสียงปรากฏการใช้ "เด้ะ" ทำให้ผู้ถอดความต้องตีความเอง ชุดข้อมูลจึงอาจยังคงมีความไม่สม่ำเสมอหากผู้ถอดความแต่ละคนตีความไม่เหมือนกัน

**หลักการถอดเป็นตัวอักษรละตินไม่ชัดเจน** แนวทางนี้ไม่ได้กำหนดการถอดคำยืมภาษาอังกฤษว่าควรใช้ตัวอักษรไทยหรืออักษรละตินอย่างแน่ชัด เนื่องจากไม่อยู่ขอบเขตของการถอดความภาษาถิ่นอีสานโดยตรง

งานในอนาคตสามารถสำรวจเพิ่มเติมได้ว่าผู้พูดส่วนใหญ่ต้องการข้อความแบบใดเมื่อคำพูดนั้นสามารถถอดความได้หลายรูปแบบ เพื่อจะพิจารณาปรับแนวทางนี้ต่อไปในอนาคต

## 5.4  ข้อจำกัดของพจนานุกรมคำอ่าน

**มีเพียงคำมูลเท่านั้น** พจนานุกรมนี้จำกัดอยู่ที่คำมูลเท่านั้น หากนำไปใช้กับงานที่ตัดคำใหญ่กว่าคำมูล จะไม่พบคำที่ใหญ่กว่าคำมูลเหล่านั้นในพจนานุกรมนี้

**มีเพียงคำที่สะกดตามแนวทางการสะกดในงานนี้เท่านั้น** พจนานุกรมนี้จำกัดเฉพาะคำที่สะกดตามแนวทางการสะกดในงานนี้เท่านั้น หากนำไปใช้กับงานที่ไม่ได้สะกดตามแนวทางดังกล่าว จะไม่พบคำที่สะกดตามแนวทางอื่นในพจนานุกรม

**คลังหน่วยเสียงอาจต่างกันไปตามกลุ่มผู้พูด** คลังหน่วยเสียงที่กำหนดขึ้นอาจจะไม่ตรงกับที่คนอีสานบางกลุ่มออกเสียงจริง เช่น ผู้พูดอีสานรุ่นใหม่อาจจะพิจารณาการออกเสียง ช เป็นเสียงพยัญชนะท้ายว่าเป็นธรรมชาติ (เช่น ออกเสียงคำว่า 'แคช' ลงท้ายด้วย

เสียง ช ช้าง ไม่ใช่เสียงแม่กดทั่วไป) ในขณะที่ผู้พูดอีสานบางกลุ่มอาจจะไม่ออกเสียงควบกล้ำเลยในตำแหน่งพยัญชนะต้น (เช่น ออกเสียงคำว่า 'เคว้งคว้าง' เป็น "เค้งค้าง")

**นิยามเกณฑ์รูปแปรหลักและรูปแปรรองยังไม่ชัดเจน** เนื่องจากยังไม่สามารถวัดได้อย่างแน่นอนว่าผู้พูดอีสานส่วนใหญ่คุ้นชินกับการออกเสียงแบบใดมากกว่า โดยเฉพาะเมื่อการออกเสียงทั้งสองแบบเป็นที่คุ้นชิน เช่น 'ฉาก' ที่สามารถออกเสียงได้ทั้ง "ฉาก" และ "สาก"

งานในอนาคตสามารถเพิ่มคำประสมและรูปแปรการสะกดที่หลากหลายมากขึ้นเพื่อให้สอดคล้องกับการใช้จริง และสามารถดัดแปลงคลังหน่วยเสียง สัทสัมผัส และเกณฑ์รูปแปรหากพบว่าผู้พูดภาษาอีสานส่วนใหญ่โน้มเอนไปทางการออกเสียงแบบอื่นใดมากกว่า

# 6 สรุป

งานนี้นำเสนอการจัดเก็บข้อมูลเสียงภาษาอีสาน โดยครอบคลุมตั้งแต่การจำแนกสำเนียงภาษาอีสาน การออกแบบการเก็บข้อมูลภาษาอีสาน มาตรฐานการสะกด หลักการถอดความ และหลักการถ่ายถอดเสียง ในแต่ละขั้นตอนของการจัดทำข้อมูลได้กำหนดหลักการขึ้นอย่างชัดเจนโดยอิงความครอบคลุม ความคุ้นชิน ความสม่ำเสมอ และหลักการทางภาษาศาสตร์

สำหรับชุดข้อมูลเสียงภาษาอีสาน พจนานุกรมคำอ่านภาษาอีสาน การจัดกลุ่มภาษาอีสาน แนวทางการสะกดคำสำหรับภาษาอีสาน แนวทางการถอดความสำหรับภาษาอีสาน และแนวทางการถ่ายถอดเสียงภาษาอีสานโดยละเอียด สามารถดูรายละเอียดเพิ่มเติมได้ที่ภาคผนวก

# รายการอ้างอิง

# กิตติกรรมประกาศ

# คณะผู้จัดทำ

- นักภาษาศาสตร์
    - อดิศัย ณ ถลาง
    - ชนกันต์ วิทยศักดิ์พันธ์
    - กฤษฎา แพทย์เจริญ
- ที่ปรึกษาด้านภาษาศาสตร์ภาษาอีสาน
    - ผศ. ดร.ศุภกิต บัวขาว　　　　　คณะมนุษยศาสตร์และสังคมศาสตร์ มหาวิทยาลัยขอนแก่น
    - ผศ. ดร.อนงค์นาฏ นุศาสตร์เลิศ　　คณะมนุษยศาสตร์และสังคมศาสตร์ มหาวิทยาลัยขอนแก่น
- ผู้ช่วยนักภาษาศาสตร์
    - กันต์ฤทัย มาสุข
    - วสวัตติ์ มาทน
    - ธนากร ก้อนสูงเนิน
    - ปฏิพล ธรรมวงษ์
    - ศักดิ์นรินทร์ พิมพ์วันคำ
    - อัมรินทร์ ชาญกล้า
    - พัฒนชัย พัสดร
- แพลตฟอร์มสำหรับงานคัดกรองเสียงและถอดความ
    - สิทธิพงศ์ ศรีไพศาลมงคล
    - ธนวิน สมุทสินธุ์
    - ชาฟัยซอล วานิ
    - จิรายุ รุ่งเรือง
    - คามิน ประกอบ
    - ปกรณ์ นาทอง
- ที่ปรึกษาโครงการ
    - ดร.วริทธิ์ ศิริโชติดำรงค์
    - ดร.พศวีร์ มานะกุล
    - อรวี สมิทธิผล
    - กฤษณพงษ์ จิรายุติ
    - สุรพล โนนสัง
    - ณัฐพงษ์ นิตราช
    - พิทวัส ทวีกิจวรชัย
    - คุณัชญ์ พิพัฒนกุล
    - กสิมะ ธารพิพิธชัย

- ฝ่ายสนับสนุนบุคลากรเพื่องานคัดกรองเสียงและถอดความ
    - แนนพวรรณ ใกล้ชิด    Looloo Technology
- ฝ่ายเก็บข้อมูลภาคสนาม
    - กฤตย์ กังวาลพงศ์พันธุ์   Wang Corporation
    - พัชรวุฒิ อ่ำเจริญ    Wang Corporation
- ฝ่ายออกแบบกราฟิก
    - ณธีพัฒน์ ดิลกเรืองพงศ์

# ภาคผนวก

**ชุดข้อมูลเสียงพูดภาษาอีสาน (Isan dialect dataset)**

    ดูเพิ่มเติมได้ที่ https://huggingface.co/datasets/scb10x/thai-isan-dialect-dataset

**พจนานุกรมคำอ่านภาษาอีสาน (Isan phonetic dictionary)**

    ดูเพิ่มเติมได้ที่ https://huggingface.co/datasets/scb10x/isan-phonetic-dictionary

**การจัดกลุ่มภาษาอีสาน (Isan dialect classification)**

    ดูเพิ่มเติมได้ที่ https://github.com/scb-10x/typhoon-isan/blob/main/isan-dialect-classification.pdf

**แนวทางการสะกดคำสำหรับภาษาอีสาน (Isan spelling standard)**

    ดูเพิ่มเติมได้ที่ https://github.com/scb-10x/typhoon-isan/blob/main/isan-spelling-standard.pdf

**แนวทางการถอดความสำหรับภาษาอีสาน (Isan speech transcription convention)**

    ดูเพิ่มเติมได้ที่ https://github.com/scb-10x/typhoon-isan/blob/main/isan-speech-transcription-convention.pdf

**แนวทางการถ่ายถอดเสียงภาษาอีสาน (Isan phonetic transcription guidelines)**

    ดูเพิ่มเติมได้ที่ https://github.com/scb-10x/typhoon-isan/blob/main/isan-phonetic-transcription-guidelines.pdf